\documentclass[letterpaper]{article} % DO NOT CHANGE THIS
\usepackage{aaai2026}  % DO NOT CHANGE THIS
\usepackage{times}  % DO NOT CHANGE THIS
\usepackage{helvet}  % DO NOT CHANGE THIS
\usepackage{courier}  % DO NOT CHANGE THIS
\usepackage[hyphens]{url}  % DO NOT CHANGE THIS
\usepackage{graphicx} % DO NOT CHANGE THIS
\usepackage{natbib}  % DO NOT CHANGE THIS AND DO NOT ADD ANY OPTIONS TO IT
\usepackage{caption} % DO NOT CHANGE THIS AND DO NOT ADD ANY OPTIONS TO IT
\usepackage{amsmath} % 提供 \operatorname 等数学命令
\usepackage{makecell}
\usepackage{bm}
\usepackage{algorithm}
\usepackage{algpseudocode}
\usepackage{booktabs}
\usepackage{amssymb}

\newcommand{\ie}{\textit{i.e.}}
\newcommand{\eg}{\textit{e.g.}}

\usepackage{newfloat}
\usepackage{listings}
\DeclareCaptionStyle{ruled}{labelfont=normalfont,labelsep=colon,strut=off} % DO NOT CHANGE THIS
\floatstyle{ruled}
\newfloat{listing}{tb}{lst}{}
\floatname{listing}{Listing}
\title{3DAlign-DAER: Dynamic Attention Policy and Efficient Retrieval Strategy for Fine-grained 3D-Text Alignment at Scale}

\author{
Yijia Fan\textsuperscript{\rm 1}\equalcontrib,
Jusheng Zhang\textsuperscript{\rm 1}\footnotemark[1],
Kaitong Cai\textsuperscript{\rm 1},
Jing Yang\textsuperscript{\rm 1},
Jian Wang\textsuperscript{\rm 2},
Keze Wang\textsuperscript{\rm 1,3}\thanks{Corresponding author: kezewang@gmail.com}
}

\affiliations{
\textsuperscript{\rm 1}Sun Yat-sen University\\
\textsuperscript{\rm 2}Snap Inc.\\
\textsuperscript{\rm 3}Guangdong Key Laboratory of Big Data Analysis and Processing\\
}

\usepackage{bibentry}
\begin{document}

\maketitle

\begin{abstract}
Despite recent advancements in 3D-text cross-modal alignment, existing state-of-the-art methods still struggle to align fine-grained textual semantics with detailed geometric structures, and their alignment performance degrades significantly when scaling to large-scale 3D databases. To overcome this limitation, we introduce \textbf{3DAlign-DAER}, a unified framework designed to align text and 3D geometry via the proposed dynamic attention policy and the efficient retrieval strategy, capturing subtle correspondences for diverse cross-modal retrieval and classification tasks. \textit{Specifically}, during the training, our proposed dynamic attention policy (DAP) employs the Hierarchical Attention Fusion (HAF) module to represent the alignment as learnable fine-grained token-to-point attentions. 
To optimize these attentions across different tasks and geometric hierarchies, our DAP further exploits the Monte Carlo tree search to dynamically calibrate HAF attention weights via a hybrid reward signal and further enhances the alignment between textual descriptions and local 3D geometry.
During the inference, our 3DAlign-DAER introduces an Efficient Retrieval Strategy (ERS) to leverage efficient hierarchical searching in the large-scale embedding spaces, outperforming traditional methods (\textit{e.g.}, KNN) in accuracy and efficiency.
\textit{Furthermore}, to facilitate text-3D alignment research and train our 3DAlign-DAER, we construct \textbf{Align3D-2M}, a large-scale dataset featuring 2M text-3D pairs, to provide sufficient fine-grained cross-modal annotations.
Extensive and comprehensive experiments demonstrate the superior performance of our 3DAlign-DAER on diverse benchmarks. Our code and updates are available at \url{https://github.com/waltstephen/3DAlign-DAER}.
\end{abstract}

\section{Introduction}

Text-3D alignment, \ie, aligning natural language descriptions with 3D geometric representations, has widespread real-world applications, such as robotic manipulation, augmented reality, and large-scale retrieval. Although existing SOTA methods~\citep{crossover,clip3d} demonstrate promising results in large-scale pre-training and global feature alignment~\citep{uni3d}, they are still faced with two critical challenges: 
\textit{i)}, existing retrieval methods often struggle at aligning fine-grained textual descriptions~\cite{crossover3, coda} (\eg, distinguishing \textit{a ceramic mug with a handle} from \textit{a simple drinking glass}) with corresponding local geometric structures (\eg, the presence or absence of a handle).
% and often struggle to precisely capture the subtle associations between textual descriptions (\eg, \textit{a wooden nightstand with a damaged left drawer}) and corresponding local geometric regions in 3D models (such as the damaged drawer). 
% Static or globally aggregated attention mechanisms \cite{attentionisalluneed} tend to overlook these critical details that determine semantic accuracy. 
This is primarily because they operate on coarse-grained feature maps or attention mechanism (\eg, global \texttt{[CLS]} token) and tend to overlook these critical geometric details that provide semantic discriminative information;
% \textit{Second}, their retrieval performance dramatically degrades as the retrieval 3D database/corpus scales up due to the inclusion of more distractors to discriminate Therefore, need an effective approach robust to sparse alignment than traditional KNN-based retrieval methods~\cite{clip} which fails
\textit{ii)}, these methods suffer from poor scalability, i.e., their performance drops sharply on larger 3D databases due to the increased difficulty of discriminating targets from challenging distractors. 
Thus, an effective retrieval method is required for robust sparse alignments to overcome the weaknesses of traditional KNN-based methods~\cite{clip} at scale. 
Besides, a dedicated, fine-grained, and large-scale text-3D alignment benchmark is missing to set the foundation for rigorous training and standardized evaluation aimed at addressing the above two challenges.
Although the existing dataset (\ie, ObjaverseXL \cite{objaverseXL}) contains scaled 3D assets, its annotated textual descriptions are mostly from noisy web sources and uncurated, which does not satisfy the need for fine-grained alignment~\cite{Z4,Z9}. \par

\begin{figure}
    \centering
    \includegraphics[width=\linewidth]{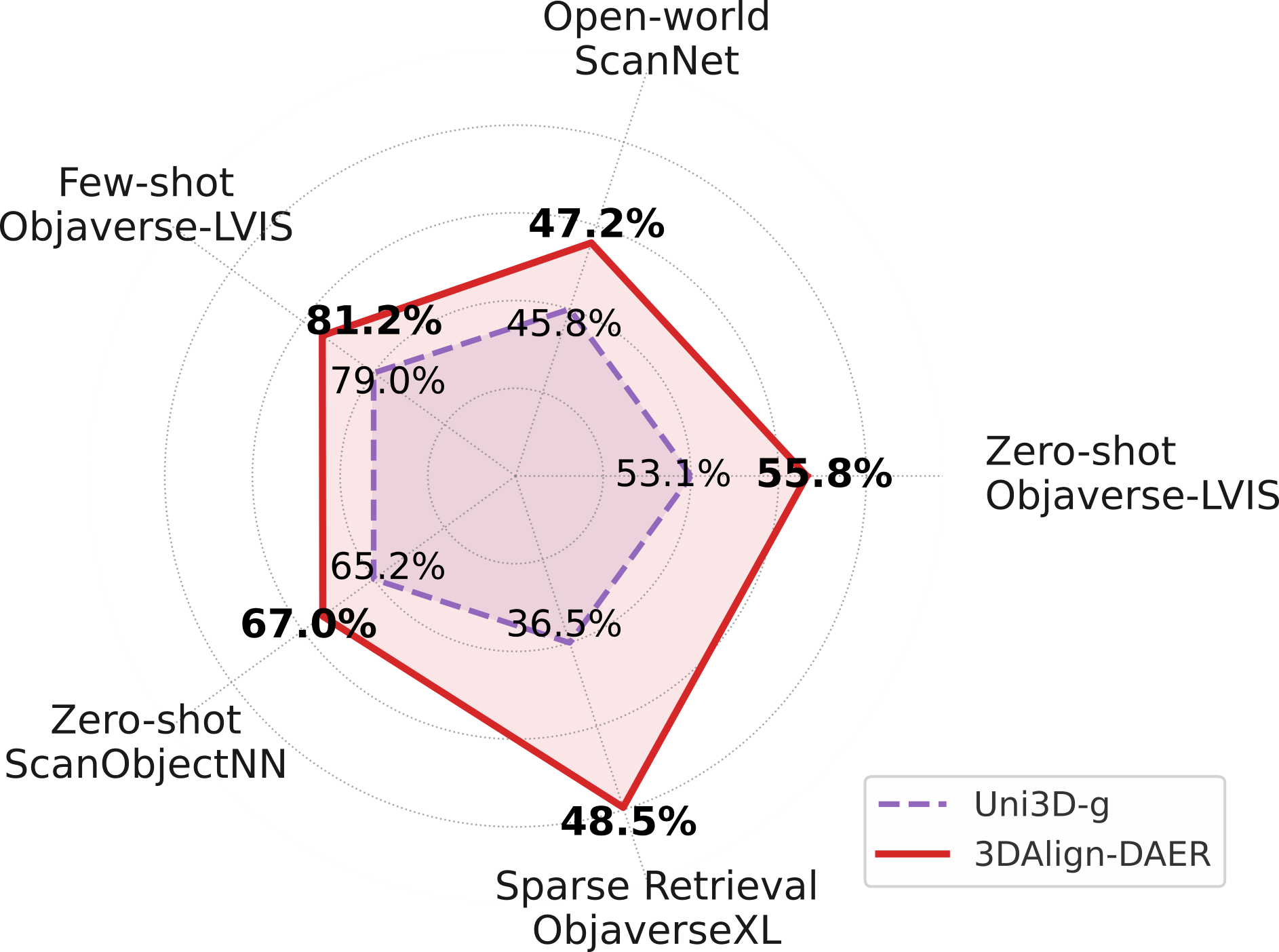}

    \caption{Our 3DAlign-DAER outperforms all task-specialized state-of-the-art methods on multiple 3D benchmarks and tasks (\ie, few/zero-shot classification, large-scale retrieval, and open-world understanding).}
    \label{fig:3DAlign_MCTS_radar}

\end{figure}

\begin{figure*}[t]
    \centering
    \includegraphics[width=0.9\linewidth]{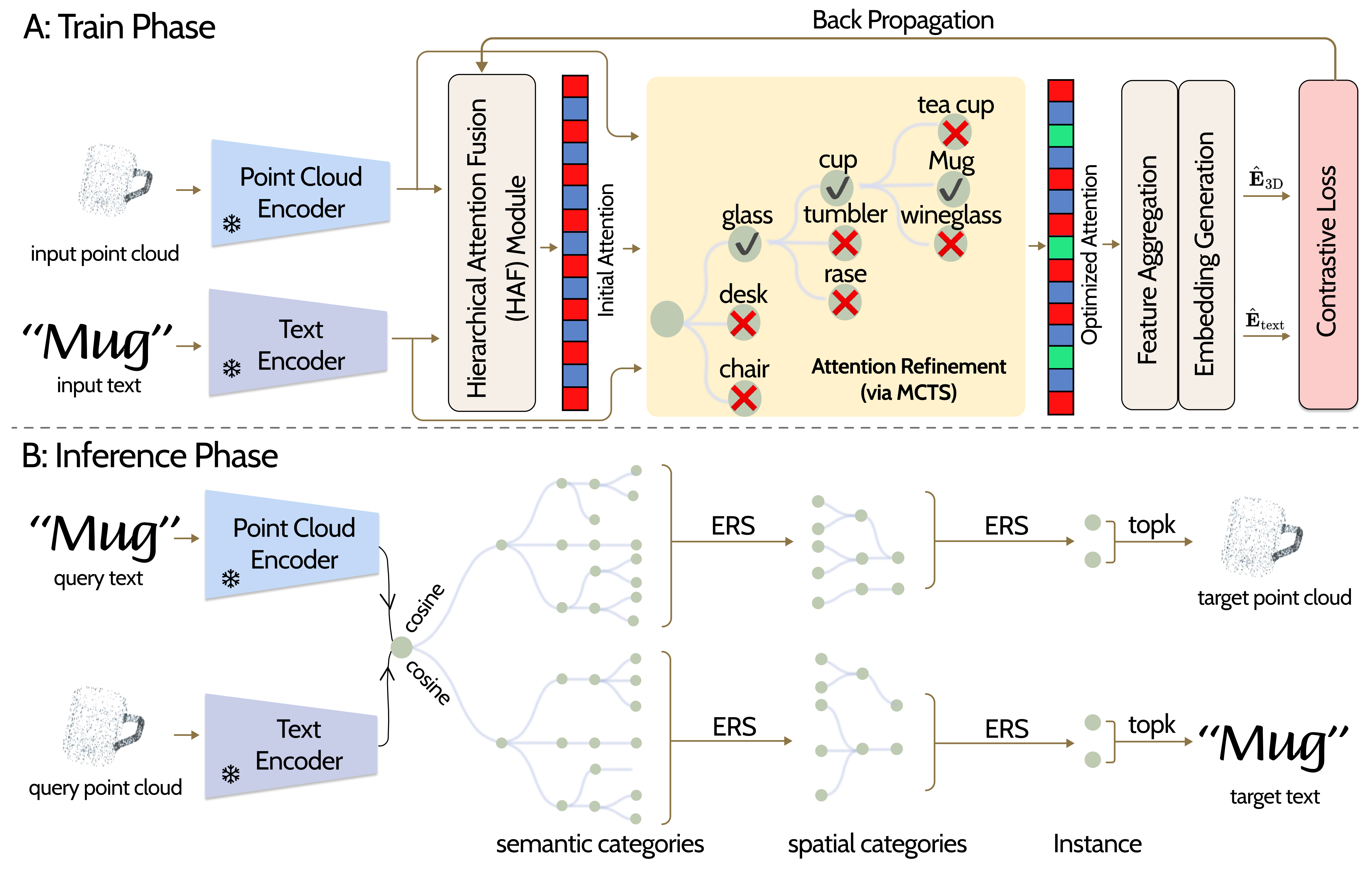}

    \caption{Overview of the 3DAlign-DAER framework. (A) Training Phase: Input modalities are initially processed by pre-trained encoders from Uni3D-g. Features are then refined through the Hierarchical Attention Fusion (HAF) module and an attention optimization module (based on MCTS) that implements the Dynamic Attention Policy (DAP). Our 3DAlign-DAER is trained using a contrastive loss on the final aggregated embeddings. (B) Inference Phase: For retrieval tasks, queries are encoded using the point cloud and text encoders belonging to the trained 3DAlign-DAER model. We propose an Efficient Retrieval Strategy (ERS) to navigate semantic and spatial categories to find the target.}
    \label{fig:pipeline}

\end{figure*}

To address these issues, we introduce our open-source dataset \textbf{Align3D-2M}. 
Constructed using our customized parallel rendering and annotation pipelines, it comprises 2 million curated text-3D pairs featuring fine-grained text-geometry correspondences. The significant scale and high quality of alignment within Align3D-2M enable the development and rigorous evaluation of models targeting complex real-world text-3D relationships for learning fine-grained alignment at scale. We further introduce our \textbf{3DAlign-DAER} framework to achieve fine-grained alignment and scalability, fully exploiting the potential of our Align3D-2M dataset. The central innovation of our 3DAlign-DAER lies in employing a Dynamic Attention Policy (DAP) to modulate the fine-grained token-to-point attentions learned by our Hierarchical Attention Fusion (HAF) module. Guided by a tailored hybrid reward from Monte Carlo Tree Search (MCTS) \cite{MCTS}, our DAP iteratively calibrates attention weights, ensuring precise correspondence between textual descriptions and local geometric details. Extensive and comprehensive results demonstrate that the alignment quality of our 3DAlign-DAER is significantly enhanced through the dynamic attention refinement of DAP. Thus, 3DAlign-DAER advances in achieving robust retrieval performance at scale by making features more discriminative against hard distractors. \par

% Specifically, in the training phase, we continue training based on Uni3D's pre-trained weights \cite{uni3d}. Instead of relying on fixed attention weights, we introduce MCTS into the Hierarchical Attention Fusion (HAF) module to perform online searches for the optimal attention modulation strategy for each training sample. Through a carefully designed reward mechanism (combining contrastive loss reduction and validation set retrieval performance), MCTS guides the model to dynamically enhance or suppress associations between specific text tokens and 3D geometric patches, thereby forcing the model to learn more precise and robust fine-grained correspondences. This approach enables the model to adaptively handle complex samples and varying semantic descriptions.

Specifically, during the training phase, our DAP employs an adaptive online search by using MCTS on attention maps from the HAF module. For each training sample, rather than relying on fixed attention weights, our DAP determines the optimal modulation strategy for the HAF weights. Our DAP is guided by our designed hybrid reward signal, which incorporates both dense feedback from the reduction in contrastive loss and sparse feedback based on retrieval performance on a validation set. By actively exploring how to enhance or suppress specific text token-to-3D patch associations based on this reward, our DAP forces the model to learn the precise and robust fine-grained correspondences necessary for challenging alignments. \par

% In the inference phase, to tackle the challenge of large-scale sparse data retrieval \cite{spare1,spare2}, we design a lightweight hierarchical retrieval algorithm called \textbf{ERS}. This algorithm utilizes knowledge learned during training to construct semantic and spatial hierarchies \cite{online1}. Through an efficient tree-based search strategy, it rapidly locates Top-K matches within the sparse embedding space, improving retrieval efficiency and accuracy on large-scale datasets and addressing the limitations of widely adopted methods (\eg, KNN, ANN, and Greedy Search).

To achieve efficient inference, especially when tackling large-scale sparse data retrieval , we design an \textbf{Efficient Retrieval Strategy (ERS)} for speed and accuracy. Our ERS leverages the highly discriminative and well-aligned representations produced by the training process of 3DAlign-DAER, which incorporates dynamic attention refinement in DAP. It constructs semantic and spatial hierarchies  over the embedding space based on the learned representations, enabling an efficient tree-based search strategy to locate Top-K matches rapidly. This approach significantly boosts retrieval performance on large-scale datasets, overcoming the speed and accuracy limitations of traditional KNN, ANN, or Greedy Search techniques. 

Comprehensive experiments show that our 3DAlign-DAER achieves SOTA on zero-shot classification, cross-modal retrieval (including large-scale), and few-shot learning, significantly outperforming strong baselines. 

Our \textbf{main} contributions are summarized as follows: i) We propose \textbf{3DAlign-DAER}, a unified framework to dynamically refine fine-grained token-to-point attention and enhance sparse text-3D alignment. Our 3DAlign-DAER further incorporates an efficient retrieval strategy for scalable and large-scale retrieval. Our 3DAlign-DAER can handle multiple geometric hierarchies and tasks; ii) We construct \textbf{Align3D-2M}, a large-scale dataset comprising $\sim$2M fine-grained text–3D pairs. We develop pipelines for parallel point cloud rendering and annotation by curating multiple noisy datasets. Align3D-2M offers rich, clean, and consistent cross-modal annotations at scale, facilitating future research for large-scale 3D-text alignment; iii) Extensive experiments across diverse benchmarks demonstrate that our 3DAlign-DAER framework achieves new SOTA performance in 3D-text alignment tasks, including zero-shot / few-shot classification and large-scale cross-modal retrieval.

\section{Related Works}

% Assuming \section{Related Work} is already defined elsewhere

\textbf{3D-Text Cross-Modal Alignment.} Learning joint representations for shapes and language often involves aligning global features using vision-language models like CLIP \cite{clip} and its variants \cite{clipvar}. Representative methods like ULIP \cite{ulip}, OpenShape \cite{liu2023openshape}, and Uni3D \cite{uni3d} leverage large datasets \cite{objaverseXL} to train 3D encoders whose features align with text embeddings in a shared space, enabling zero-shot retrieval and classification. These methods achieve strong coarse alignment by scaling models and data, sometimes enhancing text descriptions. Other works \cite{cap3d,cap3d1,scoreagg} transfer 2D knowledge by projecting 3D data or incorporate auxiliary tasks like reconstruction. However, these methods rely on aligning global representations, limiting their ability to capture fine-grained correspondences between text phrases and specific shape parts.

\textbf{Fine-Grained Alignment and Hierarchical Fusion.} Addressing the limitation of global alignment requires fine-grained grounding. While the 2D vision-language tasks using cross-modal attention \cite{crossattention,blip,Z3}, achieving this for general 3D shapes is challenging. Some efforts \cite{featurefusion,featurefusion1} explore multi-level features or apply attention in specific domains like 3D scene grounding. Yet, most existing 3D-text models \cite{3Dretrival1} lack explicit token-to-point reasoning by interacting only from a global objective. There is a requirement for hierarchical fusion methods that can link local geometry to language. Our work advances in a dynamic mechanism to achieve finer alignment.

\textbf{Search-Based Optimization for Alignment.} Standard cross-modal learning \cite{blip2,llava} relies on static architectures and gradient-based optimization. Reinforcement learning (RL) \cite{ppo,kabb} or search strategies offer potential for dynamic refinement but remain largely unexplored for 3D-text alignment. While RL has been used elsewhere in vision-language \cite{vlm-r11}, and search methods like Monte Carlo Tree Search (MCTS) \cite{MCTS,alphago} guide decisions in other domains \cite{MCTS-LLM,openai2024openaio1card,mctsphysic}, their application to directly optimize cross-modal alignment attention is novel. Inspired by MCTS, our approach actively searches the space of attention configurations during inference to discover better fine-grained alignments, departing from conventional end-to-end training or predefined dynamic routing mechanisms. 
\textbf{3D Datasets for Pretraining.} Progress has been tightly linked to dataset availability, moving from smaller annotated datasets like Text2Shape \cite{text2shape} and ShapeNet \cite{shapnet} to large web-scraped collections like Objaverse \cite{objaverse10}. These large datasets enabled the scaling of models like OpenShape \cite{liu2023openshape} and Uni3D \cite{uni3d}, improving open-vocabulary capabilities. However, large datasets often contain weak textual annotations (e.g., tags, titles), limiting fine-grained understanding despite their scale. 

 % 如果文件名有空格，建议去掉或用下划线代替
\section{Methodology}
\label{sec:method}
%designed for precise alignment between 3D point cloud representations and natural language descriptions through the integration of Monte Carlo Tree Search (MCTS).
%We first present the newly constructed dataset and illustrate the overall model architecture of our 3DAlign-DAER. %Our 3DAlign-DAER synergizes a CLIP-style text encoder with a point cloud-based 3D encoder, establishing cross-modal interaction via the Hierarchical Attention Fusion (HAF) module (\S\ref{ssec:haf}). %The core innovation of our DAP lies in employing MCTS to dynamically refine the attention weights distribution (\S\ref{ssec:mcts_modulation}). This allows our 3DAlign-DAER to adaptively adjust its focus on different spatial or semantic regions within the modalities, thereby achieving more precise inter-modal alignment (\S\ref{ssec:aggregation} and \S\ref{ssec:contrastive}). Our 3DAlign-DAER utilizes the full MCTS strategy during training to optimize the attention distribution, while the Efficient Retrieval Strategy (ERS) (\S\ref{ssec:mcts_lite}) is employed during inference for efficient retrieval.

\subsection{Align3D-2M Dataset}
\label{sec:construction} % Optional label for the subsection
To enable 3D model retrieval based on open-text instructions, we constructed a large-scale multimodal 3D dataset named Align3D-2M. During the dataset construction process, we have integrated multiple publicly available 3D model repositories, such as Objaverse-XL, ShapeNet \cite{shapnet}, Objaverse 1.0 \cite{objaverse10}, ObjectNet3D \cite{xiang2016objectnet3d}, Pix3D \cite{pix3d}, and 3D\_FUTURE \cite{3D-FUTURE}, and developed a unique data cleaning and standardization pipeline. Public repositories like ObjaverseXL, while vast, often contain raw, uncleaned metadata and include non-descriptive strings (e.g., ``a591c555d?hl=ru") unsuitable for detailed cross-modal training. Our primary goal is to create a large-scale dataset with consistent, semantically rich text annotations specifically tailored to effectively train advanced text-to-3D alignment models like our proposed 3DAlign-DAER.

To overcome the limitations of inconsistent or noisy source metadata and generate suitable training pairs, we first extract diverse 3D objects along with their available metadata (such as category labels and geometric attributes) from the integrated sources. For each object, we render a standard frontal view image. Subsequently, we utilize the multimodal capabilities of GPT-4o \cite{openai2024gpt4ocard}, providing it with both the rendered frontal image and the extracted metadata. Guided by prompt strategies incorporating this information and including randomization parameters to ensure diversity, GPT-4o is leveraged to batch-produce initial text descriptions designed for better cross-modal understanding. This process initially generated about 2 million descriptions.

To refine these descriptions, we employ a combination of automated and human reviews. Initially, a language model (a BERT-based \cite{devlin-etal-2019-bert,liu2019robertarobustlyoptimizedbert} classifier) performs preliminary screening, filtering out descriptions with obvious issues like grammatical errors, excessive generalization, or those clearly not describing 3D objects (filtering approx. 2.7\%). To verify the quality and suitability of the generated descriptions for training, we hired around 50 annotators who conducted a random sample review of about 10\% of the remaining descriptions. They focus on evaluating the accuracy, specificity, and relevance of the description to the 3D object based on predefined criteria. This multi-stage process aims to ensure a baseline level of quality and consistency across the large generated dataset, resulting in about 2 million description-object pairs deemed suitable for training our 3DAlign-DAER. Specific review criteria and statistics are detailed in the Appendix ``Details on Align3D-2M Text Annotation and Data Curation". 

\subsection{Model Architecture}
\label{ssec:architecture}

Our 3DAlign-DAER comprises four key components: a text encoder, a 3D encoder, the Hierarchical Attention Fusion (HAF) module that incorporates MCTS-driven attention modulation according to our proposed Dynamic Attention Policy (DAP), and a contrastive learning objective.

\subsubsection{Text Encoder}
\label{sssec:text_encoder}
We employ a Transformer-based architecture to process input natural language descriptions. Given an input text sequence, the encoder generates token-level features projected into a shared $d$-dimensional embedding space:
$\mathbf{F}_{\text{text}} \in \mathbb{R}^{T \times d}$, where $T$ is the number of text tokens and $d$ is the feature dimension.

\subsubsection{3D Encoder}
\label{sssec:3d_encoder}
The 3D encoder processes point cloud data, extracting geometric and spatial information. We utilize a hierarchical architecture, potentially inspired by PointNet++, involving point set sampling, local feature aggregation, and global information fusion. This transforms the 3D geometry into features compatible with the text features: $\mathbf{F}_{\text{3D}} \in \mathbb{R}^{N}$, where $N$ is the number of sampled points. Both $\mathbf{F}_{\text{text}}$ and $\mathbf{F}_{\text{3D}}$ reside in the same $d$-dimensional space.

\subsubsection{Hierarchical Attention Fusion (HAF) Module}
\label{ssec:haf} % Changed from sssec to ssec as it's a main component like MCTS modulation
The HAF module establishes fine-grained correspondences using cross-attention. Text and 3D features are linearly projected to Query ($\mathbf{Q}$), Key ($\mathbf{K}$), and Value ($\mathbf{V}$) matrices using learnable weights $\mathbf{W}_Q, \mathbf{W}_K^{\text{3D}}, \mathbf{W}_V^{\text{3D}}, \mathbf{W}_V^{\text{text}} \in \mathbb{R}^{d \times d}$. For text-to-3D attention:
\begin{align} % Using align for better spacing of multiple short equations
\label{eq:q_text}
\mathbf{Q}_{\text{text}} &= \mathbf{F}_{\text{text}} \mathbf{W}_Q & (\mathbf{Q}_{\text{text}} \in \mathbb{R}^{T \times d}) \\
\label{eq:k_3d}
\mathbf{K}_{\text{3D}} &= \mathbf{F}_{\text{3D}} \mathbf{W}_K^{\text{3D}} & (\mathbf{K}_{\text{3D}} \in \mathbb{R}^{N \times d}) \\
\label{eq:v_3d}
\mathbf{V}_{\text{3D}} &= \mathbf{F}_{\text{3D}} \mathbf{W}_V^{\text{3D}} & (\mathbf{V}_{\text{3D}} \in \mathbb{R}^{N \times d})
\end{align}
For 3D-to-text attention (used in aggregation), text values are:
\begin{equation}
\label{eq:v_text}
\mathbf{V}_{\text{text}} = \mathbf{F}_{\text{text}} \mathbf{W}_V^{\text{text}} \quad (\mathbf{V}_{\text{text}} \in \mathbb{R}^{T \times d})
\end{equation}
An initial cross-attention weight matrix is computed via scaled dot-product attention:
\begin{equation}
\label{eq:initial_attention}
\mathbf{A}_{\text{initial}} = \text{softmax}\left( \frac{\mathbf{Q}_{\text{text}} \mathbf{K}_{\text{3D}}^\top}{\sqrt{d}} \right) \in \mathbb{R}^{T \times N}
\end{equation}
The $\text{softmax}(\cdot)$ is applied row-wise. This $\mathbf{A}_{\text{initial}}$ serves as the starting point for MCTS refinement.

\subsection{MCTS-driven Dynamic Attention Policy}
\label{ssec:mcts_modulation}
To refine the initial alignment $\mathbf{A}_{\text{initial}}$, our DAP introduces MCTS to dynamically modulate the attention weights, searching for an optimized distribution $\mathbf{A}_{\text{optimized}}$ that maximizes a reward signal. 

\noindent \textbf{State ($s$):} Represents the current attention matrix $\mathbf{A} \in \mathbb{R}^{T \times N}$. The root state corresponds to $\mathbf{A}_{\text{initial}}$. \par 

\noindent \textbf{Action ($a$):} A modification operation $a: \mathbb{R}^{T \times N} \rightarrow \mathbb{R}^{T \times N}$, yielding $\mathbf{A}' = \text{Normalize}(\text{op}(\mathbf{A}, \Delta, \mathbf{M}))$. Here, $\text{op}$ is the operation (e.g., addition), $\Delta$ is the magnitude, $\mathbf{M}$ is a binary mask, and $\text{Normalize}$ re-applies row-wise softmax. Actions include enhancing or suppressing attention in $\mathbf{M}$. \par 

\noindent \textbf{Reward ($R$):} Evaluates the quality of reaching a state, incorporating a dense internal loss-oriented reward and a sparse external retrieval-oriented reward:
\begin{equation}
\label{eq:mcts_reward}
R_{\text{total}} = \alpha \cdot R_{\text{internal}} + (1 - \alpha) \cdot R_{\text{external}}
\end{equation}
% where $\alpha \in [0,1]$ balances the components.
% \begin{itemize}
%     \item $R_{\text{internal}}$: Measures relative improvement in the primary contrastive training loss (e.g., $R_{\text{internal}} = \max(0, (L_0 - L_t) / L_0)$, where $L_0$ is the loss before and $L_t$ is the loss after the MCTS state/action), reflecting optimization towards the model's core objective.
%     \item $R_{\text{external}}$: Assesses performance on the final downstream task (i.e., 3D retrieval) using metrics like weighted Recall@K and mAP. These are efficiently \textit{estimated} during MCTS rollouts to guide the search toward practical effectiveness.
% \end{itemize}
where $\alpha \in [0,1]$ is a balance factor. $R_{\text{internal}}$ denotes the decrements in the contrastive loss (see \S\ref{ssec:contrastive}) after each MCTS action. $R_{\text{external}}$ denotes the performance score of the final downstream retrieval task based on metrics (\ie, weighted Recall@K and mAP). These scores can be efficiently approximated during MCTS rollouts. 
%\subsubsection{MCTS Search Algorithm} % Elevated to subsubsection
\label{sssec:mcts_search}
MCTS iteratively builds a \textbf{search tree} via four steps: \par 

\noindent \textbf{Selection:} Traverse the tree from the root by selecting child nodes maximizing the UCT score :
\begin{equation}
\label{eq:uct}
\text{UCT}(s, a) = \overline{Q}(s, a) + c \cdot \sqrt{\frac{2 \ln N(s)}{N(s, a) + \epsilon}}
\end{equation}
where $\overline{Q}(s, a)$ is the average reward after action $a$ from state $s$, $N(s)$ and $N(s, a)$ are visit counts, $c$ is the exploration constant, and $\epsilon$ prevents division by zero. Selection stops at an expandable leaf node. \par 

\noindent \textbf{Expansion:} Add one or more child nodes corresponding to valid actions $a \in C(s)$ applicable to the leaf node's state $s$. \par 

\noindent \textbf{Simulation (Rollout):} Estimate the reward from a new node $s'$ by simulating subsequent actions using a fast policy $\pi_{\text{rollout}}$ for a fixed depth $d$: $\hat{R} = \text{Simulate}(s', \pi_{\text{rollout}}, d)$. The reward is calculated via Eq.~\ref{eq:mcts_reward}. \par 

\noindent \textbf{Backpropagation:} Propagate the estimated reward $\hat{R}$ up the path from the expanded node to the root, updating visit counts $N(s), N(s, a)$ and average rewards $\overline{Q}(s, a)$ using incremental mean updates:
\begin{align}
N(s) &\leftarrow N(s) + 1 \nonumber \\
N(s, a) &\leftarrow N(s, a) + 1 \nonumber \\
\overline{Q}(s, a) &\leftarrow \overline{Q}(s, a) + \frac{\hat{R} - \overline{Q}(s, a)}{N(s, a)} \label{eq:mcts_backprop}
\end{align}
After a fixed budget, the action corresponding to the most promising path from the root yields $\mathbf{A}_{\text{optimized}}$.

% \subsection{Feature Aggregation and Embedding Generation}
\subsection{Attention-Guided Feature Aggregation}
\label{ssec:aggregation}
The optimized attention $\mathbf{A}_{\text{optimized}}$ guides cross-modal feature aggregation:
\begin{align}
\label{eq:z_text}
\mathbf{Z}_{\text{text}} &= \mathbf{A}_{\text{optimized}} \cdot \mathbf{V}_{\text{3D}} & (\mathbf{Z}_{\text{text}} \in \mathbb{R}^{T \times d}) \\
\label{eq:z_3d}
\mathbf{Z}_{\text{3D}} &= \mathbf{A}_{\text{optimized}}^\top \cdot \mathbf{V}_{\text{text}} & (\mathbf{Z}_{\text{3D}} \in \mathbb{R}^{N \times d})
\end{align}
These features represent each modality informed by the other. Global embeddings $\mathbf{E}_{\text{text}}, \mathbf{E}_{\text{3D}} \in \mathbb{R}^{d'}$ are obtained by pooling ($\mathrm{Pool}$) and non-linear transformations ($g_{\text{text}}, g_{\text{3D}}$):

% \begin{align}
% \label{eq:e_text}
% \mathbf{E}_{\text{text}} &= g_{\text{text}}(\mathrm{Pool}(\mathbf{Z}_{\text{text}})) \\
% \label{eq:e_3d}
% \mathbf{E}_{\text{3D}} &= g_{\text{3D}}(\mathrm{Pool}(\mathbf{Z}_{\text{3D}}))
% \end{align}

% Finally, embeddings are L2-normalized:
% \begin{align}
% \label{eq:e_hat_text}
% \hat{\mathbf{E}}_{\text{text}} &= \mathbf{E}_{\text{text}} / \| \mathbf{E}_{\text{text}} \|_2 \\
% \label{eq:e_hat_3d}
% \hat{\mathbf{E}}_{\text{3D}} &= \mathbf{E}_{\text{3D}} / \| \mathbf{E}_{\text{3D}} \|_2
% \end{align}

\begin{alignat}{2}
\label{eq:e_hat_text}
\mathbf{E}_{\text{text}} &= g_{\text{text}}(\mathrm{Pool}(\mathbf{Z}_{\text{text}})) \quad & 
\mathbf{E}_{\text{3D}} &= g_{\text{3D}}(\mathrm{Pool}(\mathbf{Z}_{\text{3D}})) \\
\hat{\mathbf{E}}_{\text{text}} &= \mathbf{E}_{\text{text}} / \| \mathbf{E}_{\text{text}} \|_2 \quad & 
\hat{\mathbf{E}}_{\text{3D}} &= \mathbf{E}_{\text{3D}} / \| \mathbf{E}_{\text{3D}} \|_2
\end{alignat}

\subsection{Contrastive Learning Objective}
\label{ssec:contrastive}
Using the InfoNCE loss \cite{infonce}, the text-to-3D loss ($\mathcal{L}_{\text{t2v}}$) is for a batch of $B$ pairs:
\begin{equation}
\label{eq:infonce_t2v}
\mathcal{L}_{\text{t2v}} = -\frac{1}{B} \sum_{i=1}^{B} \log \frac{\exp\left( \hat{\mathbf{E}}_{\text{text}, i} \cdot \hat{\mathbf{E}}_{\text{3D}, i} / \tau \right)}{\sum_{j=1}^{B} \exp\left( \hat{\mathbf{E}}_{\text{text}, i} \cdot \hat{\mathbf{E}}_{\text{3D}, j} / \tau \right)}
\end{equation}
where $\tau$ is a temperature parameter. A symmetric loss $\mathcal{L}_{\text{v2t}}$ is for 3D-to-text alignment. The final bidirectional loss is:
\begin{equation}
\label{eq:infonce_bidirectional}
\mathcal{L}_{\text{bidirectional}} = \mathcal{L}_{\text{t2v}} + \mathcal{L}_{\text{v2t}}
\end{equation}
Minimizing this loss aligns positive pairs and separates negative pairs in the embedding space.

\subsection{Overall Loss Function}
\label{ssec:overall_loss}
Our overall training objective in Phase 2 is a bidirectional cross-modality contrastive loss computed with MCTS-refined attentions, with additional regularization:
\begin{equation}
\label{eq:total_loss}
\mathcal{L}_{\text{total}} = \mathcal{L}_{\text{bidirectional}} + \gamma \cdot \mathcal{L}_{\text{reg}}
\end{equation}
where $\mathcal{L}_{\text{bidirectional}}$ uses embeddings derived from $\mathbf{A}_{\text{optimized}}$, $\mathcal{L}_{\text{reg}}$ includes regularization terms (e.g., weight decay) weighted by $\gamma$. The MCTS reward (Eq.~\ref{eq:mcts_reward}) guides the \textit{search} for the optimal attention matrix $\mathbf{A}_{\text{optimized}}$, which subsequently influences $\mathcal{L}_{\text{bidirectional}}$, rather than directly contributing to the gradient via backpropagation in this formulation.

\subsection{Training Algorithm}
\label{ssec:training}
We introduce our two-stage training algorithm of 3DAlign-DAER (details in Appendix), which contains two phases: \par 

% \noindent \textbf{1$^{st}$-Phase: Contrastive Pre-training}: The model is  using $\mathcal{L}_{\text{bidirectional}}$ based on the initial attention $\mathbf{A}_{\text{initial}}$ (Eq.~\ref{eq:initial_attention}). This establishes a baseline alignment, potentially fine-tuning encoders and training projection heads. \par 
\noindent \textbf{1$^{st}$-Phase:} \textit{Contrastive Pre-training}. We pretrain only with $\mathcal{L}_{\text{bidirectional}}$ to warmup and obtain the initial attention $\mathbf{A}_{\text{initial}}$ (Eq.~\ref{eq:initial_attention}). This establishes a baseline alignment, potentially fine-tuning encoders and training projection heads. \par 

\noindent \textbf{2$^{nd}$-Phase:} \textit{MCTS-Guided Optimization}. MCTS is introduced to refine attention. The training loop involves: (1) Forward pass for $\mathbf{F}_{\text{text}}, \mathbf{F}_{\text{3D}}$. (2) Compute $\mathbf{A}_{\text{initial}}$. (3) Run MCTS search starting from $\mathbf{A}_{\text{initial}}$ to find $\mathbf{A}_{\text{optimized}}$, using the current model for reward evaluation. (4) Generate final embeddings $\hat{\mathbf{E}}_{\text{text}}, \hat{\mathbf{E}}_{\text{3D}}$ using $\mathbf{A}_{\text{optimized}}$. (5) Compute $\mathcal{L}_{\text{bidirectional}}$ (Eq.~\ref{eq:infonce_bidirectional}). (6) Backpropagate the loss to update model parameters. (7) Update MCTS statistics. MCTS search is performed periodically (e.g., every $k=10$ steps).

\subsection{ERS for Efficient Inference}
\label{ssec:mcts_lite}
During inference, the full MCTS is computationally expensive. We propose ERS, a lightweight hierarchical search strategy for efficient retrieval. It navigates a pre-computed or dynamically built hierarchical index over the 3D database. When retrieving for a query text embedding $\mathbf{q} = \hat{\mathbf{E}}_{\text{text}}$, ERS selects child nodes $a$ (representing sub-category $s_a$) using a modified UCT-like score:
\begin{equation}
\label{eq:uct_lite}
\begin{split}
&\text{UCT}_{\text{Lite}}(s, a) = \lambda_1 \cdot \text{sim}(\mathbf{q}, \bm{\mu}_{s_a}) \\
&+ \lambda_2 \cdot \frac{N_{\text{success}}(s, a)}{N(s, a) + \epsilon} + \lambda_3 \cdot \sqrt{\frac{\ln N(s)}{N(s, a) + \epsilon}}
\end{split}
\end{equation}
where $\text{sim}(\mathbf{q}, \bm{\mu}_{s_a})$ is the cosine similarity to the child's representative embedding $\bm{\mu}_{s_a}$, $N_{\text{success}}$ counts past retrieval successes through $a$, $N$ are visit counts, $\lambda_1, \lambda_2, \lambda_3$ are weights balancing similarity, historical success, and exploration, and $\epsilon$ prevents division by zero. This efficiently prunes the search space.

\section{Experiments}
We conduct a comprehensive evaluation of our 3DAlign-DAER across several key tasks. We further investigate the quality and generalization ability of the learned 3D representations. Furthermore, detailed ablation studies analyzing the contribution of different components and strategies (including MCTS optimization and reward functions), evaluations on open-world understanding (ScanNet), hyperparameter sensitivity analysis, and computational overhead assessments for both training and inference are provided in the Appendix. All experiments are performed on one A100 GPU.

\subsection{Zero-shot Shape Classification}

\subsubsection{Experimental Setup}
We select three commonly used benchmarks: Objaverse-LVIS \cite{objaverse10}, ModelNet40 \cite{ModelNet40}, and ScanObjectNN \cite{ScanObjectNN}. For comparison models, we choose ULIP-2 \cite{ulip2}, ULIP, Uni3D-B, Uni3D-L, Uni3D-g, PointCLIP \cite{Pointclip}, ReCon++-L \cite{qi2024shapellm}, and OpenShape-PointBERT \cite{liu2023openshape} to compare against our model, with each model using its officially provided inference hyperparameter configuration. We follow the evaluation setup of OpenShape, sampling 10,000 points for point clouds from Objaverse-LVIS and ModelNet40. For ScanObjectNN, we use 2,048 colorless sampled points from the OBJ ONLY version, maintaining alignment with the configuration used for testing the baseline Uni3d model. Our model is pre-trained on Align-2M and uses the previously defined ERS search during inference. Specific hyperparameters can be found in Appendix ``Implementation Details and Hyperparameters".

% Table 1: Zero-Shot 3D Classification Results
\begin{table*}[h] % [h] suggests placing the table "here" if possible
\centering
% Updated, longer caption highlighting 3DAlign-DAER performance

% Add an 'l' for the new Venue column after the first 'l' for Method
\footnotesize
\begin{tabular}{l l rrr rrr rrr}
\toprule
{Method} & {Venue} & \multicolumn{3}{c}{Objaverse-LVIS (\%)} & \multicolumn{3}{c}{ModelNet40 (\%)} & \multicolumn{3}{c}{ScanObjectNN (\%)} \\
\cmidrule(lr){3-5} \cmidrule(lr){6-8} \cmidrule(lr){9-11} % Adjust cmidrule ranges
                      &         & Top1 & Top3 & Top5 & Top1 & Top3 & Top5 & Top1 & Top3 & Top5 \\
\midrule
PointCLIP & CVPR 2022  &  1.8 &  4.0 &  5.9 & 19.4 & 28.5 & 34.9 & 10.4 & 20.9 & 30.5 \\
ULIP & CVPR 2023  &  6.1 & 13.5 & 17.8 & 60.5 & 79.1 & 84.5 & 51.4 & 71.0 & 80.3 \\
OpenShape-PointBERT & NeurIPS 2023 & 46.7 & 69.0 & 77.1 & 84.5 & 96.4 & 98.1 & 52.1 & 79.6 & 88.8 \\
ULIP-2 & CVPR 2024  & 26.7 & 44.9 & 52.5 & 75.0 & 88.0 & 93.1 & 51.7 & 72.6 & 82.4 \\
Uni3D-B & ICLR 2024   & 51.6 & 74.0 & 80.9 & 86.2 & 96.6 & 97.8 & 63.7 & 81.7 & 90.3 \\
Uni3D-L & ICLR 2024   & 53.0 & 47.2 & 81.4 & 86.4 & 96.7 & 98.2 & 64.1 & 81.9 & 90.5 \\
Uni3D-g & ICLR 2024 &  54.2  &  76.1  & 81.9  & 86.7  & 96.9  &  99.1  &  65.2 &  82.9  & 91.3  \\
ReCon++-L (shapeLLM)  & ECCV 2024 & 53.1 & 75.2 & 81.6 & 86.4 & 94.8 & 95.9 & 63.5 & 80.1 & 90.5 \\
\midrule % Optional rule to separate baseline/prior work from ours
\textbf{3DAlign-DAER (Ours)} &  & \textbf{55.8} & \textbf{77.0} & \textbf{83.1} & \textbf{88.5} & \textbf{98.6} & \textbf{99.5} & \textbf{67.0} & \textbf{86.2} & \textbf{93.1} \\
\bottomrule
\end{tabular}
\caption{Zero-Shot 3D Classification performance comparison on Objaverse-LVIS, ModelNet40, and ScanObjectNN benchmarks.}
\label{tab:zero_shot_class}
\end{table*}

\subsubsection{Experimental Results} As shown in Table \ref{tab:zero_shot_class}, our 3DAlign-DAER achieves a Top-1 accuracy of 55.8\% on the Objaverse-LVIS dataset, significantly outperforming the second-best model, ReCon++-L, at 53.1\%. On the ModelNet40 dataset, 3DAlign-DAER obtains a Top-1 accuracy of 88.5\%, also leading all comparison models. On the ScanObjectNN dataset, 3DAlign-DAER reaches a Top-1 accuracy of 67.0\%, showing a more pronounced advantage compared to Uni3D-g's 65.2\% and ReCon++-L's 63.5\%. Not only does 3DAlign-DAER lead in Top-1 accuracy, but it also comprehensively surpasses all comparison models in Top-3 and Top-5 accuracy metrics. 

\subsection{Cross-Modal Retrieval}
\subsubsection{Experimental Setup}
We conduct standard text-shape bidirectional retrieval experiments on the Text2Shape \cite{text2shape} dataset, including Shape-to-Text (S2T) and Text-to-Shape (T2S) retrieval. We adopt standard retrieval evaluation metrics: RR@1, RR@5, and NDCG@5. We select representative methods including Text2Shape, Y2Seq2Seq \cite{han2018y2seq2seqcrossmodalrepresentationlearning}, TriCoLo \cite{ruan2023tricolotrimodalcontrastiveloss}, Parts2Words \cite{tang2023parts2wordslearningjointembedding}, COM3D \cite{3dretrival2}, SCA3D \cite{sca3d}, as well as Uni3D-B, Uni3D-L and Uni3D-g. Our 3DAlign-DAER utilizes the unified embedding space learned from pre-training on Align-2M and employs the ERS to efficiently match queries (text or shape) with items in the target gallery through hierarchical search to perform the cross-modal retrieval task. Specific hyperparameter settings can be found in the Appendix. 

\subsubsection{Experimental Results}

\begin{table}[t]
  \centering

  \begingroup
    % 1. 为了让10pt的大表格能放下，建议把列间距(tabcolsep)再调小一点，例如 3pt
    \setlength{\tabcolsep}{3pt}
    
    % 2. 设置 10pt 字号 (行距 12pt) 和 罗马字体 (\rmfamily)
    \fontsize{7pt}{12pt}\selectfont

    % 注意：如果表格原本宽度超过 0.96\linewidth，adjustbox 会强行缩小表格，
    % 导致实际打印出来的字号小于 10pt。

      \begin{tabular}{lcccccc}
        \toprule
        {Method} & \multicolumn{3}{c}{S2T} & \multicolumn{3}{c}{T2S} \\
        \cmidrule(lr){2-4} \cmidrule(lr){5-7}
                 & RR@1 & RR@5 & NDCG@5 & RR@1 & RR@5 & NDCG@5 \\
        \midrule
        Text2Shape           &  0.83 &  3.37 &  0.73 &  0.40 &  2.37 &  1.35 \\
        Y2Seq2Seq            &  6.77 & 19.30 &  5.30 &  2.93 &  9.23 &  6.05 \\
        TriCoLo              & 16.33 & 45.52 & 12.73 & 10.25 & 29.07 & 19.85 \\
        Parts2Words          & 19.38 & 47.17 & 15.30 & 12.72 & 32.98 & 23.13 \\
        COM3D                & 20.03 & 48.32 & 15.62 & 13.12 & 33.48 & 23.89 \\
        Uni3D-B              & 22.51 & 51.39 & 17.05 & 14.57 & 36.28 & 25.04 \\
        Uni3D-L              & 25.02 & 53.55 & 18.71 & 15.21 & 37.57 & 27.13 \\
        Uni3D-g              & 26.23 & 54.71 & 19.42 & 16.51 & 38.32 & 28.06 \\
        SCA3D                & 27.22 & 55.56 & 19.04 & 16.67 & 38.90 & 28.17 \\
        \textbf{3DAlign-DAER} & \textbf{28.11} & \textbf{56.58} & \textbf{19.93} &
                                         \textbf{17.53} & \textbf{39.88} & \textbf{29.06} \\
        \bottomrule
      \end{tabular}

  \endgroup
    \caption{Comparison of cross-modal retrieval performance.}
  \label{tab:text2shape_retrieval_nocol}
\end{table}

As shown in Table~\ref{tab:text2shape_retrieval_nocol}, our 3DAlign-DAER achieves new state-of-the-art performance on the Text2Shape cross-modal retrieval task. It obtains the highest RR@1 scores for both Shape-to-Text (S2T) at 28.11\% and Text-to-Shape (T2S) at 17.53\%. This performance surpasses prior works, strong Uni3D baselines, and notably improves upon the previous leading method, SCA3D, by approximately 0.9\% absolute RR@1 gain in both directions, demonstrating the learning of highly discriminative joint embeddings.
The outstanding performance of 3DAlign-DAER on the cross-modal retrieval task stems from the fine-grained cross-modal alignment achieved through its DAP for dynamic attention modulation. Compared to baseline models like Uni3D-g/L, our DAP in 3DAlign-DAER can delve deeper into mining and optimizing the subtle correspondences between text and local features of 3D shapes, thereby learning joint embeddings that are richer in semantic information and more sensitive to detailed features. 

\subsection{Attention Heatmap Visualization Comparison}

\begin{figure}[t]
    \centering
    \includegraphics[width=\linewidth]{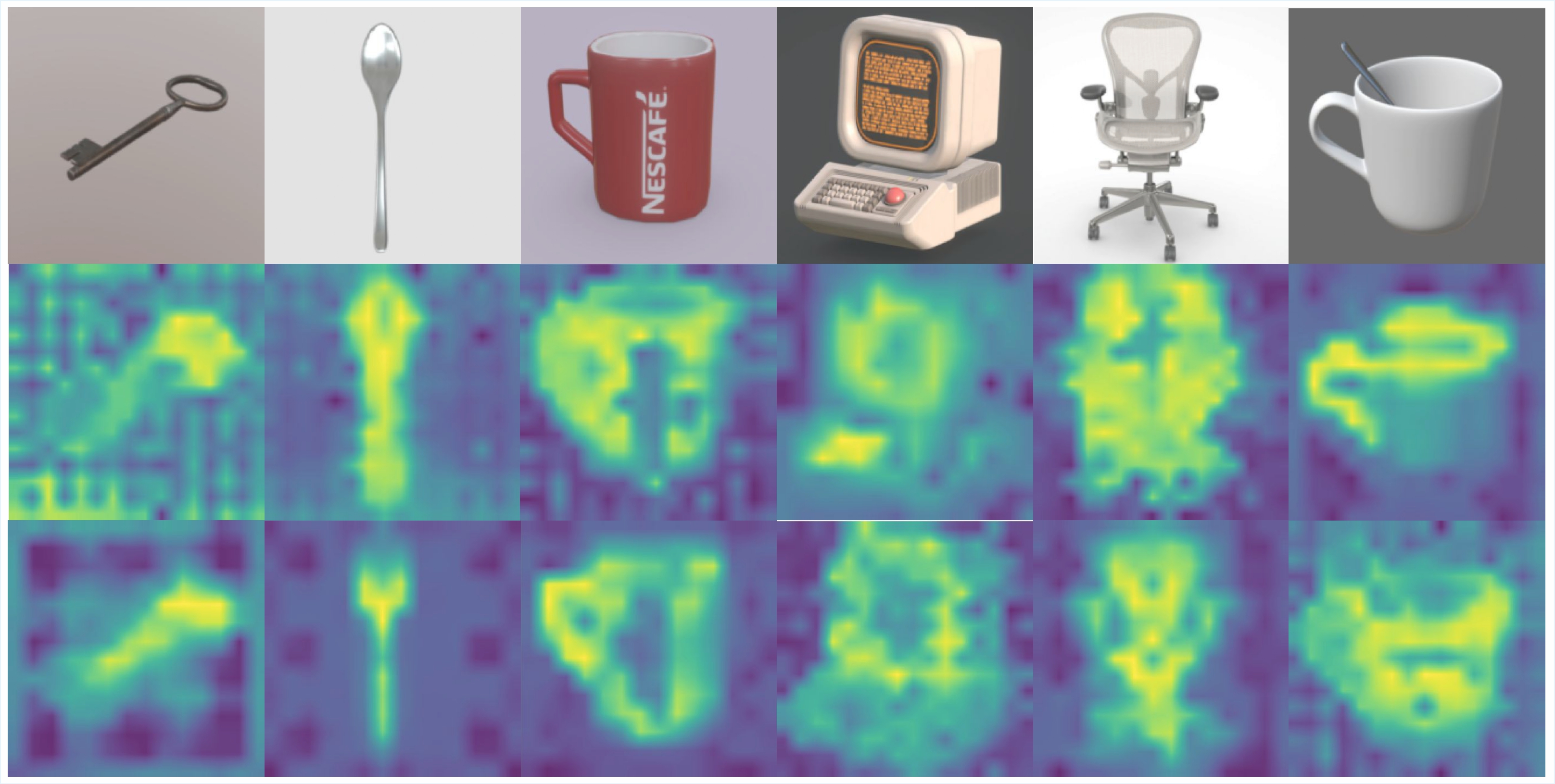}

    \caption{Attention Heatmap Visualization Comparison. Top: Original 3D objects. Middle: Attention heatmaps from Uni3D. Bottom: Attention heatmaps from 3DAlign-DAER.}
    \label{fig:attenmap}

\end{figure}

To visually verify the effectiveness of our 3DAlign-DAER in focusing attention, we perform an attention map visualization analysis, which projects the attention weights generated by the model when processing 3D point cloud data onto a 2D plane, generating intuitive heatmaps. Comparing the heatmaps generated by our 3DAlign-DAER with those from baseline models like Uni3D (see Figure \ref{fig:attenmap}), one can observe that 3DAlign-DAER's attention activations are significantly more concentrated and precise, accurately focusing on the core semantic regions of the objects. For example, for a cup, attention is focused on the cup body and handle. For a chair, attention is focused on its contours and structure. This justifies that our DAP and MCTS optimization can achieve more refined and robust fine-grained alignment.

\subsection{Large-Scale Text-to-3D Retrieval on ObjaverseXL}

\subsubsection{Experimental Setup}
To evaluate the performance of 3DAlign-DAER with its efficient retrieval strategy via ERS against several established baselines on a realistic, large-scale text-to-3D retrieval task, we utilize a subset of 1 million diverse 3D models from ObjaverseXL. Our 3DAlign-DAER is compared against Uni3D \cite{uni3d} and OpenDlign \cite{mao2024opendlign} via standard k-Nearest Neighbors (KNN) retrieval. Furthermore, to benchmark against SOTA approximate search techniques, we include comparisons with leading ANN libraries, i.e., FAISS-IVF \cite{douze2025faisslibrary}, FAISS-HNSW \cite{douze2025faisslibrary}, and DiskANN \cite{diskann}, which apply to the embeddings generated by the Uni3D baseline. We use the standard retrieval metrics, i.e., Recall@1 (R@1), Recall@5 (R@5), and Recall@10 (R@10). 

\begin{table}[t]
  \centering
  \begingroup
    \setlength{\tabcolsep}{3pt}      % 适度压缩列间距；可调
    \footnotesize                    % 缩小一档字号；可改 small/scriptsize

      \begin{tabular}{lccc}
        \toprule
        Method & R@1 (\%) & R@5 (\%) & R@10 (\%) \\
        \midrule
        Uni3D-g + KNN          &  4.1 &  9.8 & 14.5 \\
        OpenDlign + KNN        &  3.8 &  9.2 & 13.9 \\
        \midrule
        Uni3D-g + FAISS-IVF    & 33.2 & 55.4 & 65.8 \\
        Uni3D-g + FAISS-HNSW   & 35.8 & 58.1 & 68.2 \\
        Uni3D-g + DiskANN      & 36.5 & 59.0 & 69.1 \\
        \midrule
        \textbf{3DAlign-DAER + ERS} & \textbf{48.5} & \textbf{69.2} & \textbf{78.6} \\
        \bottomrule
      \end{tabular}
        \caption{Text-to-3D retrieval performance on the ObjaverseXL 1M subset.
           3DAlign-DAER with ERS surpasses traditional methods and strong ANN
           baselines.}
  \label{tab:objaverse_retrieval}
  \endgroup
\end{table}
% ========= 结束 =========
% -----------------------------------------

\subsubsection{Experimental Results}

Table~\ref{tab:objaverse_retrieval} reports text-to-3D retrieval results on the 1M-scale ObjaverseXL. Traditional KNN—used in Uni3D and OpenDlign—performs poorly (R@1 < 5\%) due to its reliance on local proximity. Strong ANN methods (FAISS-IVF/HNSW, DiskANN) improve R@1 to 33.2--36.5\%, but remain limited by the same locality assumption. In contrast, 3DAlign-DAER with ERS achieves significant gains, reaching \textbf{48.5\%} R@1, \textbf{69.2\%} R@5, and \textbf{78.6\%} R@10. This large margin demonstrates that ERS exploits hierarchical, semantically guided exploration of the embedding space—beyond purely local neighbor search—enabling accurate retrieval of globally relevant candidates in the diverse 1M-scale ObjaverseXL dataset.

\subsection{Few-shot Linear Probing}

\subsubsection{Experimental Setup}
To assess few-shot performance, we follow the standard linear-probing protocol: the pre-trained 3DAlign-DAER encoder is frozen, and only a linear classifier is trained on limited labeled samples. Experiments are conducted on Objaverse-LVIS with 1, 2, 4, 8, and 16 shots per class. We compare against representative baselines—including PointCLIP V2, ULIP, OpenShape (SparseConv and PointBERT), Uni3D, and the recent ReCon++-L (shapeLLM). All results are averaged over 10 random seeds for stability.

\subsubsection{Experimental Results}

\begin{table}[t]
  \centering
  % \resizebox{\linewidth}{!}{...} 将表格强制缩放至行宽
  \resizebox{\linewidth}{!}{
    \begin{tabular}{lccccc}
      \toprule
      \textbf{Method} & \textbf{1-s} & \textbf{2-s} & \textbf{4-s} & \textbf{8-s} & \textbf{16-s} \\
      \midrule
      PointCLIP V2 & 14.0 & 18.0 & 22.0 & 25.0 & 28.0 \\
      ULIP & 6.0 & 11.0 & 20.0 & 26.0 & 32.0 \\
      ULIP Retrained & 20.0 & 27.0 & 35.0 & 41.0 & 45.0 \\
      OpenShape-SparseConv & 22.0 & 28.0 & 36.0 & 43.0 & 48.0 \\
      OpenShape-PointBERT & 26.0 & 33.0 & 41.0 & 46.0 & 50.0 \\
      Uni3D-g & 42.0 & 54.0 & 64.0 & 74.0 & 79.0 \\
      ReCon++-L (shapeLLM) & 43.0 & 55.0 & 65.0 & 75.0 & 80.0 \\
      \textbf{3DAlign-DAER (Ours)} & \textbf{47.0} & \textbf{60.0} & \textbf{69.0} & \textbf{78.0} & \textbf{82.0} \\
      \bottomrule
    \end{tabular}
  }
    \caption{Few-shot Linear Probing on Objaverse-LVIS (Top-1 Accuracy \%)}
  \label{tab:results}
\end{table}

% ----------------------------------------------------

Table \ref{tab:results} compares 3DAlign-DAER with several advanced baselines, including ReCon++-L (shapeLLM), on few-shot linear probing (Top-1 accuracy). Across all settings (1, 2, 4, 8, 16 shots), 3DAlign-DAER achieves the best performance, consistently surpassing all competitors. The advantage is most pronounced in extremely data-scarce cases (1–2 shots), highlighting the high intrinsic quality, discriminability, and strong low-data generalization of the learned 3D representations—even when only a simple linear classifier is trained.
\section{Conclusion}
In this paper, we presented the 3DAlign-DAER framework for fine-grained alignment and large-scale sparse data retrieval in 3D-text cross-modal understanding. Besides, we construct and release the large-scale dataset Align3D-2M, containing 2 million high-quality sample pairs. Extensive and comprehensive experiments have validated the superiority of our 3DAlign-DAER and particularly demonstrated the significant advantages of our ERS in sparse retrieval scenarios. Our future work will focus on enhancing search efficiency and exploring broader applications.

\section*{Acknowledgments}
This work was supported in part by the National Natural Science Foundation of China (NSFC) under Grant 62276283, in part by the China Meteorological Administration's Science and Technology Project under Grant CMAJBGS202517, in part by Guangdong Basic and Applied Basic Research Foundation under Grant 2023A1515012985, in part by Guangdong-Hong Kong-Macao Greater Bay Area Meteorological Technology Collaborative Research Project under Grant GHMA2024Z04, in part by Fundamental Research Funds for the Central Universities, Sun Yat-sen University under Grant 23hytd006, and in part by Guangdong Provincial High-Level Young Talent Program under Grant RL2024-151-2-11.

% --- 参考文献 ---
% 1. 指定参考文献的格式样式。
%    因为你的样式文件叫 aaai2026.bst，所以这里就用 aaai2026。
\bibstyle{aaai2026}

% 2. 指定 .bib 文件的文件名（注意：没有 .bib 后缀！）。
\bibliography{AnonymousSubmission/LaTeX/aaai2026}

@String{Computing = "Computing" }

@String{Computer = "{IEEE} Computer" }

@String{Springer = "Springer-Verlag" }

@ArtifactSoftware{R,
    title = {R: A Language and Environment for Statistical Computing},
    author = {{R Core Team}},
    organization = {R Foundation for Statistical Computing},
    address = {Vienna, Austria},
    year = {2019},
    url = {https://www.R-project.org/},
}

@misc{crossover3,
      title={Enhanced Cross-modal 3D Retrieval via Tri-modal Reconstruction}, 
      author={Junlong Ren and Hao Wang},
      year={2025},
      eprint={2504.01476},
      archivePrefix={arXiv},
      primaryClass={cs.CV},
      url={https://arxiv.org/abs/2504.01476}, 
}

@inproceedings{uni3d,
  title={Uni3d: Exploring unified 3d representation at scale},
  author={Zhou, Junsheng and Wang, Jinsheng and Ma, Baorui and Liu, Yu-Shen and Huang, Tiejun and Wang, Xinlong},
  booktitle={International Conference on Learning Representations (ICLR)},
  year={2024}
}

@inproceedings{coda,
  title={CoDA: Collaborative Novel Box Discovery and Cross-modal Alignment for Open-vocabulary 3D Object Detection},
  author={Cao, Yang and Zeng, Yihan and Xu, Hang  and  Xu, Dan},
  booktitle={NeurIPS},
  year={2023}
}

@misc{clip,
      title={Learning Transferable Visual Models From Natural Language Supervision}, 
      author={Alec Radford and Jong Wook Kim and Chris Hallacy and Aditya Ramesh and Gabriel Goh and Sandhini Agarwal and Girish Sastry and Amanda Askell and Pamela Mishkin and Jack Clark and Gretchen Krueger and Ilya Sutskever},
      year={2021},
      eprint={2103.00020},
      archivePrefix={arXiv},
      primaryClass={cs.CV},
      url={https://arxiv.org/abs/2103.00020}, 
}

@article{MCTS,
   title={Monte Carlo Tree Search: a review of recent modifications and applications},
   volume={56},
   ISSN={1573-7462},
   url={http://dx.doi.org/10.1007/s10462-022-10228-y},
   DOI={10.1007/s10462-022-10228-y},
   number={3},
   journal={Artificial Intelligence Review},
   publisher={Springer Science and Business Media LLC},
   author={Świechowski, Maciej and Godlewski, Konrad and Sawicki, Bartosz and Mańdziuk, Jacek},
   year={2022},
   month=jul, pages={2497–2562} }

@ARTICLE{alphago,
  title    = "Mastering the game of Go without human knowledge",
  author   = "Silver, David and Schrittwieser, Julian and Simonyan, Karen and
              Antonoglou, Ioannis and Huang, Aja and Guez, Arthur and Hubert,
              Thomas and Baker, Lucas and Lai, Matthew and Bolton, Adrian and
              Chen, Yutian and Lillicrap, Timothy and Hui, Fan and Sifre,
              Laurent and van den Driessche, George and Graepel, Thore and
              Hassabis, Demis",
  journal  = "Nature",
  volume   =  550,
  number   =  7676,
  pages    = "354--359",
  month    =  oct,
  year     =  2017
}

@inproceedings{crossover,
author={Sayan Deb Sarkar and Ondrej Miksik and Marc Pollefeys and Daniel Barath and Iro Armeni},
title={CrossOver: 3D Scene Cross-Modal Alignment}, 
booktitle = {The IEEE Conference on Computer Vision and Pattern Recognition (CVPR)},
year = {2025}
}

@article{clipvar,
  title={SCLIP: Rethinking Self-Attention for Dense Vision-Language Inference},
  author={Wang, Feng and Mei, Jieru and Yuille, Alan},
  journal={arXiv preprint arXiv:2312.01597},
  year={2023}
}

@inproceedings{ulip,
  title={Ulip: Learning a unified representation of language, images, and point clouds for 3d understanding},
  author={Xue, Le and Gao, Mingfei and Xing, Chen and Mart{\'\i}n-Mart{\'\i}n, Roberto and Wu, Jiajun and Xiong, Caiming and Xu, Ran and Niebles, Juan Carlos and Savarese, Silvio},
  booktitle={CVPR},
  pages={1179--1189},
  year={2023}
}

@misc{liu2023openshape,
      title={OpenShape: Scaling Up 3D Shape Representation Towards Open-World Understanding}, 
      author={Minghua Liu and Ruoxi Shi and Kaiming Kuang and Yinhao Zhu and Xuanlin Li and Shizhong Han and Hong Cai and Fatih Porikli and Hao Su},
      year={2023},
      eprint={2305.10764},
      archivePrefix={arXiv},
      primaryClass={cs.CV}
}

@article{objaverseXL,
  title={Objaverse-XL: A Universe of 10M+ 3D Objects},
  author={Matt Deitke and Ruoshi Liu and Matthew Wallingford et al.},
  journal={arXiv preprint arXiv:2307.05663},
  year={2023}
}

@article{cap3d,
      title={Scalable 3D Captioning with Pretrained Models},
      author={Luo, Tiange and Rockwell, Chris and Lee, Honglak and Johnson, Justin},
      journal={arXiv preprint arXiv:2306.07279},
      year={2023}
}

@article{cap3d1,
      title={View Selection for 3D Captioning via Diffusion Ranking},
      author={Luo, Tiange and Johnson, Justin and Lee, Honglak},
      journal={arXiv preprint arXiv:2404.07984},
      year={2024}
}

@inproceedings{scoreagg,
 author = {Kabra, Rishabh and Matthey, Loic and Lerchner, Alexander and Mitra, Niloy J.},
 title = {Leveraging VLM-Based Pipelines to Annotate 3D Objects},
 booktitle = {ICML},
 series = {Proceedings of Machine Learning Research},
 publisher = {PMLR},
 volume = {235},
 year = {2024}
}

@misc{crossattention,
      title={Cross-Modal Self-Attention Network for Referring Image Segmentation}, 
      author={Linwei Ye and Mrigank Rochan and Zhi Liu and Yang Wang},
      year={2019},
      eprint={1904.04745},
      archivePrefix={arXiv},
      primaryClass={cs.CV},
      url={https://arxiv.org/abs/1904.04745}, 
}

@misc{featurefusion,
      title={Multi-scale Feature Fusion with Point Pyramid for 3D Object Detection}, 
      author={Weihao Lu and Dezong Zhao and Cristiano Premebida and Li Zhang and Wenjing Zhao and Daxin Tian},
      year={2024},
      eprint={2409.04601},
      archivePrefix={arXiv},
      primaryClass={cs.CV},
      url={https://arxiv.org/abs/2409.04601}, 
}

@misc{featurefusion1,
      title={Deep Hough Voting for 3D Object Detection in Point Clouds}, 
      author={Charles R. Qi and Or Litany and Kaiming He and Leonidas J. Guibas},
      year={2019},
      eprint={1904.09664},
      archivePrefix={arXiv},
      primaryClass={cs.CV},
      url={https://arxiv.org/abs/1904.09664}, 
}

@article{3Dretrival1,
title = {TextANIMAR: Text-based 3D animal fine-grained retrieval},
journal = {Computers and Graphics},
volume = {116},
pages = {162-172},
year = {2023},
issn = {0097-8493},
doi = {https://doi.org/10.1016/j.cag.2023.07.026},
url = {https://www.sciencedirect.com/science/article/pii/S0097849323001553},
author = {Trung-Nghia Le and Tam V. Nguyen et al.},
}

@INPROCEEDINGS{3dretrival2,
  author={Wu, Hao and Li, Ruochong and Wang, Hao and Xiong, Hui},
  booktitle={ICME}, 
  title={COM3D: Leveraging Cross-View Correspondence and Cross-Modal Mining for 3D Retrieval}, 
  year={2024},
  volume={},
  number={},
  pages={1-6},
  doi={10.1109/ICME57554.2024.10688095}
}

@misc{blip2,
      title={BLIP-2: Bootstrapping Language-Image Pre-training with Frozen Image Encoders and Large Language Models}, 
      author={Junnan Li and Dongxu Li and Silvio Savarese and Steven Hoi},
      year={2023},
      eprint={2301.12597},
      archivePrefix={arXiv},
      primaryClass={cs.CV},
      url={https://arxiv.org/abs/2301.12597}, 
}

@misc{llava,
      title={Visual Instruction Tuning}, 
      author={Haotian Liu and Chunyuan Li and Qingyang Wu and Yong Jae Lee},
      year={2023},
      eprint={2304.08485},
      archivePrefix={arXiv},
      primaryClass={cs.CV},
      url={https://arxiv.org/abs/2304.08485}, 
}

@misc{vlm-r11,
  author       = {Shen, Haozhan and Zhang, Zilun and Zhao, Kangjia and Zhang, Qianqian and Xu, Ruochen and Zhao, Tiancheng},
  title        = {VLM-R1: A stable and generalizable R1-style Large Vision-Language Model},
  howpublished = {\url{https://github.com/om-ai-lab/VLM-R1}},
  note         = {Accessed: 2025-02-15},
  year         = {2025}
}

@misc{openai2024openaio1card,
      title={OpenAI o1 System Card}, 
      author={OpenAI},
      year={2024},
      eprint={2412.16720},
      archivePrefix={arXiv},
      primaryClass={cs.AI},
      url={https://arxiv.org/abs/2412.16720}, 
}

@misc{MCTS-LLM,
      title={ReST-MCTS*: LLM Self-Training via Process Reward Guided Tree Search}, 
      author={Dan Zhang and Sining Zhoubian and Ziniu Hu and Yisong Yue and Yuxiao Dong and Jie Tang},
      year={2024},
      eprint={2406.03816},
      archivePrefix={arXiv},
      primaryClass={cs.CL},
      url={https://arxiv.org/abs/2406.03816}, 
}

@misc{text2shape,
      title={Text2Shape: Generating Shapes from Natural Language by Learning Joint Embeddings}, 
      author={Kevin Chen and Christopher B. Choy and Manolis Savva and Angel X. Chang and Thomas Funkhouser and Silvio Savarese},
      year={2018},
      eprint={1803.08495},
      archivePrefix={arXiv},
      primaryClass={cs.CV},
      url={https://arxiv.org/abs/1803.08495}, 
}

@misc{shapnet,
      title={ShapeNet: An Information-Rich 3D Model Repository}, 
      author={Angel X. Chang and Thomas Funkhouser and Leonidas Guibas and Pat Hanrahan and Qixing Huang and Zimo Li and Silvio Savarese and Manolis Savva and Shuran Song and Hao Su and Jianxiong Xiao and Li Yi and Fisher Yu},
      year={2015},
      eprint={1512.03012},
      archivePrefix={arXiv},
      primaryClass={cs.GR},
      url={https://arxiv.org/abs/1512.03012}, 
}

@article{objaverse10,
  title={Objaverse: A Universe of Annotated 3D Objects},
  author={Matt Deitke and Dustin Schwenk and Jordi Salvador and Luca Weihs and
          Oscar Michel and Eli VanderBilt and Ludwig Schmidt and
          Kiana Ehsani and Aniruddha Kembhavi and Ali Farhadi},
  journal={arXiv preprint arXiv:2212.08051},
  year={2022}
}

@inproceedings{pix3d,
  title={Pix3D: Dataset and Methods for Single-Image 3D Shape Modeling},
  author={Sun, Xingyuan and Wu, Jiajun and Zhang, Xiuming and Zhang, Zhoutong and Zhang, Chengkai and Xue, Tianfan and Tenenbaum, Joshua B and Freeman, William T},
  booktitle={CVPR},
  year={2018}
}

@misc{3D-FUTURE,
      title={3D-FUTURE: 3D Furniture shape with TextURE}, 
      author={Huan Fu and Rongfei Jia and Lin Gao and Mingming Gong and Binqiang Zhao and Steve Maybank and Dacheng Tao},
      year={2020},
      eprint={2009.09633},
      archivePrefix={arXiv},
      primaryClass={cs.CV},
      url={https://arxiv.org/abs/2009.09633}, 
}

@incollection{xiang2016objectnet3d,
    author = {Xiang, Yu and Kim, Wonhui and Chen, Wei and Ji, Jingwei and Choy, Christopher 
           and Su, Hao and Mottaghi, Roozbeh and Guibas, Leonidas and Savarese, Silvio},
    title = {ObjectNet3D: A Large Scale Database for 3D Object Recognition},
    booktitle = {ECCV},
    year = {2016}
}

@misc{openai2024gpt4ocard,
      title={GPT-4o System Card}, 
      author={OpenAI},
      year={2024},
      eprint={2410.21276},
      archivePrefix={arXiv},
      primaryClass={cs.CL},
      url={https://arxiv.org/abs/2410.21276}, 
}

@misc{liu2019robertarobustlyoptimizedbert,
      title={RoBERTa: A Robustly Optimized BERT Pretraining Approach}, 
      author={Yinhan Liu and Myle Ott and Naman Goyal and Jingfei Du and Mandar Joshi and Danqi Chen and Omer Levy and Mike Lewis and Luke Zettlemoyer and Veselin Stoyanov},
      year={2019},
      eprint={1907.11692},
      archivePrefix={arXiv},
      primaryClass={cs.CL},
      url={https://arxiv.org/abs/1907.11692}, 
}

@inproceedings{devlin-etal-2019-bert,
    title = "{BERT}: Pre-training of Deep Bidirectional Transformers for Language Understanding",
    author = "Devlin, Jacob  and
      Chang, Ming-Wei  and
      Lee, Kenton  and
      Toutanova, Kristina",
    editor = "Burstein, Jill  and
      Doran, Christy  and
      Solorio, Thamar",
    booktitle = "ACL",
    month = jun,
    year = "2019",
    address = "Minneapolis, Minnesota",
    publisher = "Association for Computational Linguistics",
    url = "https://aclanthology.org/N19-1423/",
    doi = "10.18653/v1/N19-1423",
    pages = "4171--4186",
}

@INPROCEEDINGS {ModelNet40,
author = { Zhirong Wu and Song, Shuran and Khosla, Aditya and Fisher Yu and Linguang Zhang and Xiaoou Tang and Xiao, Jianxiong },
booktitle = {CVPR},
title = {{ 3D ShapeNets: A deep representation for volumetric shapes }},
year = {2015},
volume = {},
ISSN = {1063-6919},
pages = {1912-1920},
doi = {10.1109/CVPR.2015.7298801},
url = {https://doi.ieeecomputersociety.org/10.1109/CVPR.2015.7298801},
publisher = {IEEE Computer Society},
address = {Los Alamitos, CA, USA},
month =Jun
}

@inproceedings{ScanObjectNN,
      title = {Revisiting Point Cloud Classification: A New Benchmark Dataset and Classification Model on Real-World Data},
      author = {Mikaela Angelina Uy and Quang-Hieu Pham and Binh-Son Hua and Duc Thanh Nguyen and Sai-Kit Yeung},
      booktitle = {ICCV},
      year = {2019}
}

@misc{ulip2,
      title={ULIP-2: Towards Scalable Multimodal Pre-training for 3D Understanding}, 
      author={Le Xue and Ning Yu and Shu Zhang and Artemis Panagopoulou and Junnan Li and Roberto Martín-Martín and Jiajun Wu and Caiming Xiong and Ran Xu and Juan Carlos Niebles and Silvio Savarese},
      year={2024},
      eprint={2305.08275},
      archivePrefix={arXiv},
      primaryClass={cs.CV},
      url={https://arxiv.org/abs/2305.08275}, 
}

@article{Pointclip,
  title={PointCLIP: Point Cloud Understanding by CLIP},
  author={Zhang, Renrui and Guo, Ziyu and Zhang, Wei and Li, Kunchang and Miao, Xupeng and Cui, Bin and Qiao, Yu and Gao, Peng and Li, Hongsheng},
  journal={arXiv preprint arXiv:2112.02413},
  year={2021}
}

@article{qi2024shapellm,
  author = {Qi, Zekun and Dong, Runpei and Zhang, Shaochen and Geng, Haoran and Han, Chunrui and Ge, Zheng and Yi, Li and Ma, Kaisheng},
  title = {ShapeLLM: Universal 3D Object Understanding for Embodied Interaction},
  journal = {arXiv preprint arXiv:2402.17766},
  year = {2024}
}

@misc{han2018y2seq2seqcrossmodalrepresentationlearning,
      title={Y2Seq2Seq: Cross-Modal Representation Learning for 3D Shape and Text by Joint Reconstruction and Prediction of View and Word Sequences}, 
      author={Zhizhong Han and Mingyang Shang and Xiyang Wang and Yu-Shen Liu and Matthias Zwicker},
      year={2018},
      eprint={1811.02745},
      archivePrefix={arXiv},
      primaryClass={cs.CV},
      url={https://arxiv.org/abs/1811.02745}, 
}

@misc{ruan2023tricolotrimodalcontrastiveloss,
      title={TriCoLo: Trimodal Contrastive Loss for Text to Shape Retrieval}, 
      author={Yue Ruan and Han-Hung Lee and Yiming Zhang and Ke Zhang and Angel X. Chang},
      year={2023},
      eprint={2201.07366},
      archivePrefix={arXiv},
      primaryClass={cs.CV},
      url={https://arxiv.org/abs/2201.07366}, 
}

@misc{tang2023parts2wordslearningjointembedding,
      title={Parts2Words: Learning Joint Embedding of Point Clouds and Texts by Bidirectional Matching between Parts and Words}, 
      author={Chuan Tang and Xi Yang and Bojian Wu and Zhizhong Han and Yi Chang},
      year={2023},
      eprint={2107.01872},
      archivePrefix={arXiv},
      primaryClass={cs.CV},
      url={https://arxiv.org/abs/2107.01872}, 
}

@article{sca3d,
  title={SCA3D: Enhancing Cross-modal 3D Retrieval via 3D Shape and Caption Paired Data Augmentation},
  author={Ren, Junlong and Wu, Hao and Xiong, Hui and Wang, Hao},
  journal={arXiv preprint arXiv:2502.19128},
  year={2025}
}

@article{mao2024opendlign,
  title={OpenDlign: Enhancing Open-World 3D Learning with Depth-Aligned Images},
  author={Mao, Ye and Jing, Junpeng and Mikolajczyk, Krystian},
  journal={arXiv preprint arXiv:2404.16538},
  year={2024}
}

@misc{douze2025faisslibrary,
      title={The Faiss library}, 
      author={Matthijs Douze and Alexandr Guzhva and Chengqi Deng and Jeff Johnson and Gergely Szilvasy and Pierre-Emmanuel Mazaré and Maria Lomeli and Lucas Hosseini and Hervé Jégou},
      year={2025},
      eprint={2401.08281},
      archivePrefix={arXiv},
      primaryClass={cs.LG},
      url={https://arxiv.org/abs/2401.08281}, 
}

@inproceedings{diskann,
 author = {Jayaram Subramanya, Suhas and Devvrit, Fnu and Simhadri, Harsha Vardhan and Krishnawamy, Ravishankar and Kadekodi, Rohan},
 booktitle = {NeurIPS},
 editor = {H. Wallach and H. Larochelle and A. Beygelzimer and F. d\textquotesingle Alch\'{e}-Buc and E. Fox and R. Garnett},
 pages = {},
 publisher = {Curran Associates, Inc.},
 title = {DiskANN: Fast Accurate Billion-point Nearest Neighbor Search on a Single Node},
 url = {https://proceedings.neurips.cc/paper_files/paper/2019/file/09853c7fb1d3f8ee67a61b6bf4a7f8e6-Paper.pdf},
 volume = {32},
 year = {2019}
}

@article{clip3d,
 title={CLIP goes 3D: Leveraging Prompt Tuning for Language Grounded 3D Recognition},
 author={Hegde, Deepti and Valanarasu, Jeya Maria Jose and Patel, Vishal M},
 journal={arXiv preprint arXiv:2303.11313},
 year={2023}
}

@inproceedings{blip,
      title={BLIP: Bootstrapping Language-Image Pre-training for Unified Vision-Language Understanding and Generation}, 
      author={Junnan Li and Dongxu Li and Caiming Xiong and Steven Hoi},
      year={2022},
      booktitle={ICML},
}

@misc{ppo,
      title={Proximal Policy Optimization Algorithms}, 
      author={John Schulman and Filip Wolski and Prafulla Dhariwal and Alec Radford and Oleg Klimov},
      year={2017},
      eprint={1707.06347},
      archivePrefix={arXiv},
      primaryClass={cs.LG},
      url={https://arxiv.org/abs/1707.06347}, 
}

@misc{kabb,
      title={KABB: Knowledge-Aware Bayesian Bandits for Dynamic Expert Coordination in Multi-Agent Systems}, 
      author={Jusheng Zhang and Zimeng Huang and Yijia Fan and Ningyuan Liu and Mingyan Li and Zhuojie Yang and Jiawei Yao and Jian Wang and Keze Wang},
      year={2025},
      eprint={2502.07350},
      archivePrefix={arXiv},
      primaryClass={cs.AI},
      url={https://arxiv.org/abs/2502.07350}, 
}

@misc{mctsphysic,
      title={Symbolic Physics Learner: Discovering governing equations via Monte Carlo tree search}, 
      author={Fangzheng Sun and Yang Liu and Jian-Xun Wang and Hao Sun},
      year={2023},
      eprint={2205.13134},
      archivePrefix={arXiv},
      primaryClass={cs.AI},
      url={https://arxiv.org/abs/2205.13134}, 
}

@misc{infonce,
      title={Representation Learning with Contrastive Predictive Coding}, 
      author={Aaron van den Oord and Yazhe Li and Oriol Vinyals},
      year={2019},
      eprint={1807.03748},
      archivePrefix={arXiv},
      primaryClass={cs.LG},
      url={https://arxiv.org/abs/1807.03748}, 
}

@inproceedings{
Z3,
title={{CF}-{VLM}: CounterFactual Vision-Language Fine-tuning},
author={Jusheng Zhang and Kaitong Cai and Yijia Fan and Jian Wang and Keze Wang},
booktitle={NeurIPS},
year={2025},
url={https://openreview.net/forum?id=0qGtaRTsCo}
}

@inproceedings{
Z4,
title={{MAT}-Agent: Adaptive Multi-Agent Training Optimization},
author={Jusheng Zhang and Kaitong Cai and Yijia Fan and Ningyuan Liu and Keze Wang},
booktitle={NeurIPS},
year={2025},
url={https://openreview.net/forum?id=YDWRTYgR79}
}

@misc{Z9,
      title={DrDiff: Dynamic Routing Diffusion with Hierarchical Attention for Breaking the Efficiency-Quality Trade-off}, 
      author={Jusheng Zhang and Yijia Fan and Kaitong Cai and Zimeng Huang and Xiaofei Sun and Jian Wang and Chengpei Tang and Keze Wang},
      year={2025},
      eprint={2509.02785},
      archivePrefix={arXiv},
      primaryClass={cs.CL},
      url={https://arxiv.org/abs/2509.02785}, 
}

% \newpage
% \newpage

% % --- 附录部分 (如果需要) ---
% \appendix

% \input{AnonymousSubmission/LaTeX/sections/Details_on_3DAlign-DAER_Training_Algorithm}
% \input{AnonymousSubmission/LaTeX/sections/Implementation_Details_and_Hyperparameters}
% \input{AnonymousSubmission/LaTeX/sections/Ablation_experiment}
% \input{AnonymousSubmission/LaTeX/sections/Open-World_Understanding}
% \input{AnonymousSubmission/LaTeX/sections/Ablation_experiment}
% \input{AnonymousSubmission/LaTeX/sections/MCTS_Training_Overhead_Analysis}
% \input{AnonymousSubmission/LaTeX/sections/ERS_Inference_Overhead_Analysis}
% \input{AnonymousSubmission/LaTeX/sections/Study_on_the_Contribution_of_Reward_Function_Components}
% \input{AnonymousSubmission/LaTeX/sections/Details_on_Align3D-2M_Tex_Annotation_and_Data_Curation}
% \input{AnonymousSubmission/LaTeX/sections/Theoretical_Analysis_of_MCTS_Computational_Overhead}
\end{document}